%% file: main.tex
\documentclass[journal]{IEEEtran}

\usepackage{amsmath, amsthm, amssymb, mathtools, bm}
\usepackage{lmodern}
\usepackage{color}
\usepackage{algorithm, algorithmic}
\usepackage{enumerate}
\usepackage{hyperref}
\usepackage{cleveref}
\usepackage{xspace}

\usepackage{tikz}
\usetikzlibrary{shapes.geometric, arrows, positioning, shapes.symbols}

\tikzstyle{state} = [rectangle,
rounded corners,
draw,
fill=gray!10,
align=center,
]

\tikzstyle{io} = [trapezium, 
trapezium stretches=true,
trapezium left angle=70, 
trapezium right angle=110, 
minimum width=2cm, 
align=center,
draw=black,
fill=blue!10
]

\tikzstyle{process} = [rectangle, 
minimum width=2cm, 
align=center,
draw=black, 
fill=orange!10
]

\tikzstyle{decision} = [diamond, 
minimum width=2cm, 
align=center,
draw=black, 
fill=green!10
]

\tikzstyle{arrow} = [-latex,
rounded corners,
thick,
->,
>=stealth
]

\newtheorem{definition}{Definition}
\newtheorem{lemma}{Lemma}
\newtheorem{theorem}{Theorem}
\newtheorem{corollary}{Corollary}

\newtheorem{remark}{Remark}

\newcommand{\w}{\text{Wealth}}

\newcommand{\br}[1]{\left(#1\right)}
\newcommand{\bigo}[1]{\mathcal{O}\br{#1}}
\newcommand{\norm}[1]{\left\lVert{#1}\right\rVert}

\DeclareMathOperator{\sech}{sech}

\newcommand{\Hil}{\mathcal{H}} 
\newcommand{\algname}{{DECO}\xspace}

\crefname{equation}{Eq.}{Eqs.}

\title{Decentralized Parameter-Free Online Learning}

\author{Tomas Ortega, \IEEEmembership{Graduate Student Member, IEEE}, and Hamid Jafarkhani, \IEEEmembership{Fellow, IEEE}
    \thanks{
        Authors are with the Center for Pervasive Communications \& Computing and  EECS Department, University of California, Irvine, Irvine, CA 92697 USA (e-mail: \{tomaso, hamidj\}@uci.edu).
This work was supported in part by the NSF Award ECCS-2207457.
}
}

\begin{document}
\maketitle

\begin{abstract}
    We propose the first parameter-free decentralized online learning algorithms with network regret guarantees, which achieve sublinear regret without requiring hyperparameter tuning.
    This family of algorithms connects multi-agent coin-betting and decentralized online learning via gossip steps.
    To enable our decentralized analysis, we introduce a novel ``betting function'' formulation for coin-betting that simplifies the multi-agent regret analysis.
    Our analysis shows sublinear network regret bounds and is validated through experiments on synthetic and real datasets.
    This family of algorithms is applicable to distributed sensing, decentralized optimization, and collaborative ML applications.
\end{abstract}

\section{Introduction}
\IEEEPARstart{O}{nline} learning algorithms are essential for adapting to streaming data and changing environments, such as in online routing, ad selection, and spam filtering~\cite{hazan_introduction_2023,shalev-shwartz_online_2011}.
In the standard Online Convex Optimization (OCO) framework, a learner makes a decision $x_{t}$ from a convex set $\mathcal{X}$ at each round $t$.
It then incurs a loss $l_{t}(x_{t})$, determined by a convex loss function $l_t$ that is unknown in advance.
The goal is to minimize the \textbf{regret} after $T$ rounds, which measures the cumulative loss difference between the learner's decisions and those of a single decision in hindsight
\begin{equation}
    R_{T}(u) := \sum_{t=1}^{T} l_{t}(x_t) - \sum_{t=1}^{T} l_{t}(u),
\end{equation}
where $u \in \mathcal{X}$ is an arbitrary comparator point.

\subsubsection*{The challenge of parameter tuning}
Classic algorithms like Online Gradient Descent (OGD) achieve optimal regret, but this performance hinges on careful \textbf{hyperparameter tuning}~\cite{hazan_introduction_2023}.
Typically, an algorithm's learning rate must be set based on prior knowledge of problem parameters, such as the total time horizon $T$, bounds on the comparator's norm $\norm{u}$, or properties of the loss functions.
This dependency is a significant practical drawback.
In many real-world applications, these parameters are unknown, or the environment is non-stationary, making a single, fixed learning rate suboptimal.
Incorrect tuning can lead to poor performance, requiring costly trial-and-error to find a suitable setting~\cite{orabona_coin_2016}.

\subsubsection*{Parameter-free learning}
To overcome this limitation, \textbf{parameter-free algorithms} have been developed~\cite{mcmahan_no-regret_2012,mcmahan_unconstrained_2014,orabona_coin_2016,cutkosky_online_2017}.
These methods are based on an analogy to \textbf{coin-betting}, where minimizing regret is reframed as maximizing a gambler's wealth in a sequential betting game.
These algorithms achieve excellent bounds (which are optimal in some settings) without requiring any hyperparameter tuning.
They automatically adapt to the characteristics of the problem, giving regret guarantees that depend on the unknown comparator $u$, hence why they are also known as comparator-adaptive algorithms~\cite{van_der_hoeven_comparator-adaptive_2020, hoeven_distributed_2021}.
When the decision set $\mathcal{X}$ is unbounded and losses are $L$-Lipschitz, it has been shown that no algorithm can guarantee a regret better than $\Omega\br{\norm{u}L \sqrt{T \ln(\norm{u} LT + 1)}}$.
Parameter-free algorithms can achieve this bound~\cite{mcmahan_no-regret_2012}.

\subsubsection*{The decentralized setting}
While parameter-free methods solve the tuning problem for a single learner, many modern applications require collaboration among multiple agents in a network.
For example, today's large-scale systems, from sensor networks to collaborative machine learning, are inherently \textbf{decentralized}~\cite{nedic_rate_2007,johansson_simple_2007}.
In these systems, a network of $N$ agents must collaborate to solve a global learning task without a central coordinator.
Each agent $n$ makes its own decisions $x_{n,t}$, incurs a local loss $l_{n,t}(x_{n,t})$, and can only communicate with its immediate neighbors over a network graph $\mathcal{G}$.
The collective goal is to minimize the \textbf{network regret}\footnote{We use the nomenclature from~\cite{achddou_distributed_2024}, but similar notions have been previously called \emph{collective} regret~\cite{hsieh_multi-agent_2022}.}
\begin{equation}
    R_{T}^{\text{net}}(u) := \sum_{t=1}^{T} \frac{1}{N} \sum_{n=1}^{N} l_{t}(x_{n,t}) - \sum_{t=1}^{T} l_{t}(u), \label{eq:network_regret}
\end{equation}
where $l_t(x) = \frac{1}{N}\sum_{m=1}^{N} l_{m,t}(x)$ is the global loss function.
This formulation measures the performance of the entire network's decisions against a single best-in-hindsight comparator $u$.

There have been recent efforts to solve variants of this problem in graphs where agents are available intermittently~\cite{achddou_distributed_2024}, or where a single agent is activated per round~\cite{hoeven_distributed_2021}, but to the best of our knowledge, no parameter-free decentralized approach with \emph{network regret guarantees} has been proposed to date.
Note that the decentralized formulation described above also encompasses distributed scenarios with a central coordinator and various computing nodes.
This is simply a special case of the topology of $\mathcal{G}$.

\subsubsection*{Contributions}
This paper introduces the first decentralized online learning algorithms that are parameter-free and come with network regret guarantees.
We bridge the gap between adaptive single-agent learning and multi-agent coordination by combining the coin-betting framework with decentralized gossip protocols.

Our main contributions are:
\begin{itemize}
    \item We propose a novel family of decentralized algorithms that allow agents to collaboratively minimize cumulative loss without any hyperparameter tuning.
    \item We provide theoretical analysis showing that our algorithms achieve sublinear regret, comparable to centralized methods.
    \item We introduce a new \emph{betting function} formulation that simplifies the analysis of coin-betting algorithms in the multi-agent context.
\end{itemize}
We also validate our approach with experiments on synthetic and real-world datasets, demonstrating its effectiveness in decentralized networks.

The remainder of the paper is organized as follows.~\Cref{sec:algorithm} introduces our proposed algorithm and its underlying intuition.
\Cref{sec:preliminaries} provides the necessary technical background for our analysis.
\Cref{sec:analysis} presents the main theoretical results and proofs.
\Cref{sec:operational-details} provides operational details to ease implementation.
Finally,~\Cref{sec:experiments} discusses our experimental findings, and~\Cref{sec:conclusion} concludes the paper.

\section{The decentralized coin-betting algorithm}\label{sec:algorithm}
In this section, we present our decentralized parameter-free algorithm.
We begin with a high-level and intuitive explanation of its mechanics before providing the formal algorithm description and performance analysis.

\subsection{High-Level idea: local betting and global gossiping}
The main idea behind our approach is that each agent in the network plays a \textbf{personalized betting game}.
The goal is to maximize wealth, which, through the lens of online learning theory, is equivalent to minimizing regret.

At each round, every agent $n$ uses its accumulated knowledge of past losses (summarized in a vector $G_{n,t-1}$) to decide on its next action, $x_{n,t}$.
This decision is analogous to a gambler \textbf{placing a bet}.
The size and direction of the bet are determined by a parameter-free strategy derived from the coin-betting framework, which automatically adapts to observed data without needing a learning rate.

After making a decision, each agent observes a local subgradient $g_{n,t}$ of its loss function.
This subgradient acts as the ``outcome'' of the betting round, informing the agent how good its bet was.
The agent then updates its internal state (its accumulated knowledge $G_{n,t}$) based on this new information.

If agents only acted on their local information, they might never achieve a globally good solution.
To promote consensus, agents communicate with their neighbors after each round.
They engage in a \textbf{gossip protocol}, where they exchange and average their internal states.
This process spreads information across the network, ensuring that over time, all agents' decisions are guided by the collective knowledge of the entire system.

The core of our contribution lies in exploiting the best properties of both tools.
The coin-betting mechanism provides local, adaptive, and parameter-free decisions, while the gossip protocol provides a decentralized mechanism for global consensus.

\subsection{Algorithm description}
Inspired by the coin-betting aspect, we name our algorithms \algname{}, for Decentralized Coin-betting.
The formal procedure, performed at each agent, is detailed in~\Cref{alg:decentralized-coin-betting}.
Each agent $n$ maintains a state vector $G_{n,t}$, which accumulates the subgradients it has observed, updated by the gossip process.
At the start of each round $t$, the agent uses this state to compute its decision $x_{n,t}$.

We propose two closely related variants for making this decision:
\begin{itemize}
    \item \textbf{\algname{}-\pmb{i}}: This version simply adds a gossip step to the standard coin-betting approach from~\cite{orabona_coin_2016}, where the decision $x_{n,t}$ depends on both the agent's state $G_{n,t-1}$ and its current wealth, $\w_{n,t-1}$.
          While effective, tracking and gossiping wealth adds complexity to the algorithm and analysis.
    \item \textbf{\algname{}-\pmb{ii}}: This is a novel formulation, which uses a \textbf{betting function} $h_t(\cdot)$ (defined in~\Cref{sec:preliminaries}) to compute the decision $x_{n,t}$ solely from the state $G_{n,t-1}$.
          This simplifies the algorithm, as only the state vectors need to be gossiped, and makes the network regret analysis more tractable.
\end{itemize}

After receiving the subgradient $g_{n,t}$, each agent performs a local update on its state.
Then, all agents execute one or more rounds of gossip by applying a gossip matrix $W$.
This matrix is doubly stochastic and reflects the network topology, meaning agents only need to communicate with their direct neighbors.
This gossip step averages the states across neighbors, driving the entire system toward consensus.

\begin{algorithm}[t]
    \caption{Decentralized Coin-Betting (\algname{}), variants (\colorbox{red!30}{\pmb{i}}) and (\colorbox{green!30}{\pmb{ii}})}\label{alg:decentralized-coin-betting}
    \input{decentralized_algo.tex}
\end{algorithm}

\subsection{Summary of main results}
Before diving into the detailed description and mathematical proofs, we discuss our main theoretical guarantees.
\subsubsection{The average local regret is independent of the topology}
Our first key result (\Cref{thm:local_avg_regret_bound}) shows that the \emph{average} of local regrets across all agents is sublinear and matches the bounds for centralized parameter-free algorithms.
Remarkably, this guarantee holds regardless of the network's topology or connectivity.
Even if the agents never communicate ($W=I$), their average performance is still optimal.
The only necessary condition is that the gossip matrix $W$ must be doubly stochastic.

\subsubsection{Network regret and disagreement}
Our second key result (\Cref{thm:network_loss_regret_bound}) characterizes the more challenging \textbf{network regret}.
We prove that it consists of two components: the sublinear average local regret mentioned above, plus a non-negative \textbf{disagreement term}.
The latter term measures the cumulative cost incurred because agents' decisions differ from one another.
For a complete graph, $\mathcal{G}$, the disagreement term is null, recovering \emph{both} the single-agent scenario and the regret guarantees.

\subsubsection{Controlling disagreement with gossip}
Finally, we show that the disagreement term is controlled by the rate of information mixing in the network, which is determined by the spectral properties of the gossip matrix $W$ and the number of gossip steps performed.
By performing a number of gossip steps that grows slowly with time (e.g.~linearly with the round number $t$), we can ensure that the disagreement term also grows sublinearly, leading to an overall sublinear network regret (\Cref{lemma:square-root-disagreement}).
This shows that given enough communication, we can achieve sublinear network regret.

\section{Preliminaries for the analysis}\label{sec:preliminaries}
To formally prove the guarantees stated in the previous section, we now introduce the key mathematical concepts upon which our algorithm is built.
We describe each component (subgradients, coin-betting, and gossip) that plays a role in our framework separately.

\subsection{Subgradients and Online Convex Optimization}
Our analysis applies to the general problem of Online Convex Optimization (OCO).
A standard technique for analyzing OCO problems is to linearize the convex loss functions using subgradients.
This allows us to bound the regret of the convex problem by analyzing an equivalent Online Linear Optimization (OLO) problem.

\begin{definition}[Subgradient]
    For a convex function $f: \mathcal{X}\rightarrow \mathbb{R}$, a vector $g \in \mathcal{X}$ is a \emph{subgradient} of $f$ at a point $x \in \mathcal{X}$ if for all $y \in \mathcal{X}$:
    \begin{equation*}
        f(y) \geq f(x) + \langle g, y - x \rangle.
    \end{equation*}
\end{definition}
This inequality implies that the regret is upper-bounded by the sum of linear losses
\begin{equation}
    R_{T}(u) \leq \sum_{t=1}^{T} \langle g_{t}, x_{t} - u \rangle.
\end{equation}
Our algorithm uses the subgradients $g_{n,t}$ of the local loss functions as the primary feedback signal for learning.
Note that through online-to-batch conversion results~\cite{cesa-bianchi_generalization_2004}, OLO algorithms can also solve convex stochastic optimization problems.

\subsection{The coin-betting framework}
The coin-betting framework is the engine that drives the local, adaptive learning of each agent.
It provides a powerful method for designing parameter-free algorithms by establishing a duality between online learning and a sequential betting game.
The formal connection is established by the reward-regret relationship, which states that an upper bound on regret is mathematically equivalent to a lower bound on cumulative reward (or wealth in the betting analogy).

\begin{lemma}[Reward-Regret Relationship~\cite{mcmahan_unconstrained_2014}]\label{lemma:reward-regret}
    Let $V$, $V^*$ be a pair of dual vector spaces.
    Let $F: V \to \mathbb{R} \cup \{+\infty\}$ be a proper convex lower semi-continuous function and let $F^*: V^* \to \mathbb{R} \cup \{+\infty\}$ be its Fenchel conjugate.
    Let $x_1, x_2, \dots, x_T \in V$ and $g_1, g_2, \dots, g_T \in V^*$.
    For any $\epsilon > 0$,
    \begin{align}
        \underbrace{\sum_{t=1}^T \langle g_t, x_t \rangle}_{\text{Reward}} \geq F\br{-\sum_{t=1}^T g_t } - \epsilon
    \end{align}
    if and only if
    \begin{align}
        \forall u \in V^*, \quad \underbrace{\sum_{t=1}^T \langle g_t, x_t - u \rangle}_{\text{Regret}} \leq F^*(u) + \epsilon.
    \end{align}
\end{lemma}
This lemma tells us that if we can guarantee that our cumulative earnings (reward) are lower-bounded by some function $F$ of the cumulative outcomes, then our regret is automatically upper-bounded by its conjugate $F^*$.

The function $F$ is constructed using a sequence of \textbf{coin-betting potentials}.
These are carefully designed functions that represent a guaranteed lower bound on a gambler's wealth.
\begin{definition}[Coin Betting Potential~\cite{orabona_coin_2016}]\label{def:potential}
    Let $\epsilon > 0$.
    Let ${\{ F_t \}}_{t=0}^\infty$ be a sequence of functions $F_t: (-a_t, a_t) \to \mathbb{R}_+$ where $a_t > t$.
    The sequence ${\{ F_t \}}_{t=0}^\infty$ is called a \textbf{sequence of coin betting potentials for initial endowment $\epsilon$} if it satisfies the following conditions:
    \begin{enumerate}[(a)]
        \item $F_0(0) = \epsilon$.
        \item For every $t \geq 0$, $F_t(x)$ is even, logarithmically convex, strictly increasing on $[0, a_t)$, and $\lim_{x \to a_t} F_t(x) = +\infty$. 
        \item For every $t \geq 1$, every $x \in [-(t-1), t-1]$, and every $c \in [-1,1]$, it holds that
              \begin{align}
                  \br{1 + c \beta_t} F_{t-1}(x) \geq F_t(x + c),
              \end{align}
              where
              \begin{align}
                  \beta_t = \frac{F_t(x + 1) - F_t(x - 1)}{F_t(x + 1) + F_t(x - 1)}.
              \end{align}
    \end{enumerate}
    Moreover, it is an \textbf{excellent} sequence of coin betting potentials if it also satisfies
    \begin{enumerate}[(d)]
        \item For every $t \geq 0$, $F_t$ is twice-differentiable and satisfies $x \cdot F''_t(x) \geq F'_t(x)$ for every $x \in [0, a_t)$. 
    \end{enumerate}
\end{definition}
Popular choices for these potentials, such as the Krichevsky-Trofimov (KT) potential or the exponential potential (see~\Cref{tab:mappings}), lead to betting strategies that produce near-optimal regret bounds of the form $\mathcal{O}(\norm{u}\sqrt{T \ln \br{1 + T}})$.
\begin{table}[H]
    \centering
    \caption{Mappings from $F_t(x)$ to $\beta_t(x)$.}\label{tab:mappings}
    \begin{tabular}{|c|c|c|}
        \hline
        Name                & $F_t(x)$                                                                                                             & $\beta_t(x)$            \\
        \hline
        Exponential         & $\frac{\epsilon}{\sqrt{t}} \exp\br{\frac{x^2}{2t}}$                                                                  & $\tanh\br{\frac{x}{t}}$ \\
        \hline
        Krichevsky-Trofimov & $\frac{2^t \cdot \Gamma\br{\frac{t+1}{2} + \frac{x}{2}} \cdot \Gamma\br{\frac{t+1}{2} - \frac{x}{2}}}{\pi \cdot t!}$ & $\frac{x}{t}$           \\
        \hline
    \end{tabular}
\end{table}

\subsubsection*{Betting functions}
For our decentralized analysis, we introduce a new tool called a \textbf{betting function}.
This function, derived from the potentials, maps the accumulated subgradients directly to a decision (a bet).
This decouples the betting decision from the agent's wealth, simplifying the algorithm and its analysis.

\begin{definition}[Betting function]\label{def:betting-function}
    Let $\{F_t\}$ be a sequence of coin-betting potentials.
    The betting function at time $t > 0$ is defined as $h_t(x) = \beta_t(x) \cdot F_{t-1}(x)$, where $\beta_t(x)$ is the betting fraction derived from $F_t$.
\end{definition}
This function $h_t$ is used in \algname{}-\pmb{ii} to determine $x_{n,t}$ based only on the gossiped state $G_{n,t-1}$.

\subsection{Gossip algorithms for consensus}\label{subsection:consensus}
Gossip algorithms are the communication backbone of our method, enabling agents to coordinate and reach a consensus.
They are decentralized protocols for averaging information across a network.

An agent's communication with its neighbors is defined by a \textbf{gossip matrix} $W$.
\begin{definition}[Gossip Matrix]\label{def:gossip_matrix}
    Given a connected graph $\mathcal{G} = (\mathcal{V}, \mathcal{E})$, a real $N\times N$ matrix $W$ is a \textbf{gossip matrix} if it is doubly stochastic ($\sum_m W_{mn} = \sum_n W_{mn} = 1$), its non-zero entries correspond to the edges of the graph ($W_{mn}>0$ only if $(m,n) \in \mathcal{E}$ or $m=n$), and its spectral radius $\rho(W - \frac{1}{N} \mathbf {1} \mathbf{1}^\top)$ is less than 1.
\end{definition}
Crucially, such a matrix can be constructed using only local information (e.g., via the Metropolis-Hastings method~\cite{boyd_randomized_2006}), making it suitable for decentralized systems.
When a vector of $N$ values is multiplied by $W$, each agent's value becomes a weighted average of its own and its neighbors' values.
Repeated application of $W$ causes all agents' values to converge to the network-wide average.
The rate of this convergence is geometric and is determined by the spectral properties of the network graph, captured by $\rho$.
\begin{lemma}[Gossip convergence rate~\cite{boyd_randomized_2006}]\label{lemma:gossip}
    Let $W$ be a gossip matrix and let $\bar{x} = \frac{1}{N} \sum_{n=1}^N x_{n,0}$.
    After $k$ applications of $W$, the vector $x_k = W^k x_0$ converges to the average value $\bar{x} \otimes \mathbf{1}$:
    \begin{align}
        \norm{x_{k} - \bar{x} \otimes \mathbf{1}} \leq \rho^k \norm{x_0 - \bar{x} \otimes \mathbf{1}}.
    \end{align}
\end{lemma}
This exponential convergence is what allows our algorithm to efficiently control the disagreement between agents and ensure the entire network collaborates effectively on the global learning task.

For our analysis, we will need the following corollary, proved using the triangle inequality $\norm{x_{n,k} - \bar{x}} \leq  \norm{x_{k} - \bar{x} \otimes \mathbf{1}}$.
\begin{corollary}[Gossip component-wise convergence rate]\label{cor:gossip-components}
    Given the conditions in~\Cref{lemma:gossip}, the components of the gossip iterations converge to the mean as
    \begin{align}
        \norm{x_{n,k} - \bar{x}}\leq \rho^k \norm{x_0 - \bar x}.
    \end{align}
\end{corollary}
To ease the analysis, without loss of generality, we will often write that \algname{} does one round of gossip per round of coin-betting.
The reasoning is identical for $q(t)$ rounds of gossip per round of coin-betting, since the multiplication of two gossip matrices is also a gossip matrix.

\section{Analysis}\label{sec:analysis}

\subsection{Multi-agent betting game}

In this subsection, we extend the single-agent coin-betting framework to a multi-agent setting suitable for decentralized online learning.

\subsubsection*{Single-agent coin-betting recap}

In the single-agent coin-betting game~\cite{orabona_coin_2016}, a gambler starts with an initial wealth $\w_0 = \epsilon > 0$.
At each round $t$, the gambler decides how much to bet $x_t$ on the outcome of a coin flip.
The coin outcome $c_t \in [-1,1]$, which is a real number\footnote{Although the outcome is not binary, we use the terminology of flipping a coin to be consistent with the existing literature.} between -1 and 1, satisfies $c_t = -g_t$, where $g_t$ is the subgradient from the OLO formulation, assuming $g_t$ is a scalar.\footnote{We refer the reader to~\cite{orabona_coin_2016} for the extension to vectors in a Hilbert space.}
The wealth updates according to
\begin{align}
    \w_t & = \w_{t-1} + c_t x_t  \\
         & = \w_{t-1} - g_t x_t.
\end{align}
By using a betting strategy based on coin-betting potentials $F_t$ as in~\Cref{def:potential}, we can lower-bound the wealth and thereby upper-bound the regret through~\Cref{lemma:reward-regret}.

\subsubsection*{Extension to multi-agent setting}

In our multi-agent setting, each agent $n$ acts as a gambler with initial wealth $\w_{n,0} = \epsilon > 0$.
At each round $t$, each agent decides its own betting amount $x_{n,t}$ based on its local information.
After the betting decision, agents exchange information with their neighbors via a gossip protocol.

More formally, for the \emph{local betting strategy}, each agent $n$ computes its betting fraction using the coin-betting potential
\begin{equation}\label{eq:beta_nt}
    \beta_{n,t} = \frac{F_t(C_{n,t-1} + 1) - F_t(C_{n,t-1} - 1)}{F_t(C_{n,t-1} + 1) + F_t(C_{n,t-1} - 1)},
\end{equation}
and sets
\begin{equation}\label{eq:x_nt}
    x_{n,t} = \beta_{n,t} \w_{n,t-1}.
\end{equation}
Next, upon receiving the coin outcome $c_{n,t} \in [-1,1]$, each agent updates its wealth and accumulated coin output
\begin{align}
    \hat \w_{n,t} & = \w_{n,t-1} + c_{n,t} x_{n,t},\label{eq:w_nt} \\
    \hat C_{n,t}  & = C_{n,t-1} + c_{n,t}.\label{eq:C_nt}
\end{align}
Finally, agents exchange their wealth $\hat \w_{n,t}$, and accumulated coin outputs $\hat C_{n,t}$ with their neighbors.
The updated variables after gossip are
\begin{align}
    \w_{n,t} & = \sum_{m=1}^N W_{mn} \hat \w_{m,t},\label{eq:gossiped_w} \\
    C_{n,t}  & = \sum_{m=1}^N W_{mn} \hat C_{m,t}.\label{eq:gossiped_c}
\end{align}

Notice how the gossip exchange step ensures that agents' decisions and coin outputs are mixed and averaged across the network, preserving the network-wide accumulated coin output $\sum_{s=1}^t \bar c_s$, where $\bar c_t$ denotes the average coin outcome at time $t$.

\subsubsection*{Wealth lower bound in multi-agent setting}

Our goal is to establish a wealth lower bound similar to the single-agent case.
Specifically, we aim to show that
\begin{equation}\label{eq:wealth_lb}
    \w_{n,t} \geq F_t(C_{n,t}).
\end{equation}
By averaging this equation across all agents, we obtain
\begin{align}
    \overline{\w}_{t} & = \frac{1}{N} \sum_{n=1}^N \w_{n,t}                            \\
                      & \geq \frac{1}{N} \sum_{n=1}^N F_t(C_{n,t})                     \\
                      & \geq F_t \br{\frac{1}{N} \sum_{n=1}^N C_{n,t}}                 \\
                      & = F_t \br{\sum_{s=1}^t \bar c_s}, \label{eq:wealth_lb_average}
\end{align}
where we have used the log-convexity of $F_t$ and the fact that the average of the accumulated coin outputs is equal to the accumulated average coin output.

To prove~\Cref{eq:wealth_lb}, we proceed by induction.
\begin{lemma}[Wealth lower bound in multi-agent setting]\label{lemma:multi_agent_wealth_lb}
    Under the proposed multi-agent betting game and assuming the coin-betting potential $F_t$ satisfies the conditions in~\Cref{def:potential}, the wealth of each agent satisfies:
    \begin{align}
        \w_{n,t} \geq F_t(C_{n,t}),
    \end{align}
    as long as the gossip matrix $W$ is doubly stochastic.
\end{lemma}

\begin{proof}
    We use induction on $t$.
    For the base case, $t=0$, for any agent $n$, we have $\w_{n,0} = \epsilon = F_0(0)$ and $C_{n,0} = 0$, so the inequality holds.

    For the inductive step, we assume the inequality holds at time $t-1$.
    At time $t$, the wealth update at Agent $n$ is given by:
    \begin{align}
        \w_{n,t} & = \sum_{m=1}^N W_{mn} \hat \w_{m,t}                                      \\
                 & = \sum_{m=1}^N W_{mn} \br{\w_{m, t-1} + c_{m,t} x_{m,t} }                \\
                 & = \sum_{m=1}^N W_{mn} \w_{m,t-1}\br{1 + \beta_{m,t} c_{m,t}}             \\
                 & \geq \sum_{m=1}^N W_{mn} F_{t-1}(C_{m,t-1})\br{1 + \beta_{m,t} c_{m,t}}.
    \end{align}
    The last step is derived by the induction hypothesis, $\w_{m,t-1} \geq F_{t-1}(C_{m,t-1})$.
    Using the coin-betting potential property (c) in~\Cref{def:potential}, we have:
    \begin{align}\label{eq:wealth_update_bound}
        F_{t-1}(C_{m,t-1})\br{1 + \beta_{m,t} c_{m,t}} & \geq F_t(C_{m,t-1} + c_{m,t}) \\
                                                       & = F_t(\hat C_{m,t}).
    \end{align}
    Since $F_t$ is logarithmically convex and $W$ is doubly stochastic, applying Jensen's inequality yields:
    \begin{align}
        \w_{n,t} & \geq \sum_{m=1}^N W_{mn} F_t(\hat C_{m,t})   \\
                 & \geq F_t\br{\sum_{m=1}^N W_{mn}\hat C_{m,t}} \\
                 & = F_t\br{C_{n,t}}.
    \end{align}
\end{proof}
This lemma shows that each agent's wealth is lower bounded by the coin-betting potential evaluated at its accumulated coin output.
This relationship forms the basis for the regret bounds of our algorithm.
\begin{remark}
    Note that~\Cref{lemma:multi_agent_wealth_lb} also applies for a betting-function based decision.
    The proof follows identically, but is omitted for brevity.
\end{remark}

\subsection{Regret analysis}

Using~\Cref{lemma:multi_agent_wealth_lb}, we can derive regret bounds for our algorithms.
Note that we have shown its proof for the case where $\{g_t\}$ are scalars, but the proof can be extended to elements of any Hilbert space $\Hil$ as in~\cite[Theorem 3]{orabona_coin_2016}.

\begin{theorem}[Average Local Regret Bound]\label{thm:local_avg_regret_bound}
    Assuming that the coin-betting potential $F_t$ satisfies the conditions in~\Cref{def:potential}, both versions of \algname{} achieve the following regret bound for any comparator $u \in \Hil$:
    \begin{align}
        R_T^{\text{local}}(u) & := \frac{1}{N} \sum_{n=1}^N \sum_{t=1}^T l_{n,t}(x_{n,t}) - l_{n,t}(u) \\
                              & \leq F_T^*(u) + \epsilon,
    \end{align}
    where $F_T^*$ is the Fenchel conjugate of $F_T$, and $\epsilon > 0$ is the initial endowment.
\end{theorem}
\begin{proof}
    Notice that, due to convexity, we have:
    \begin{align*}
        \frac{1}{N} \sum_{n=1}^N \sum_{t=1}^T l_{n,t}(x_{n,t}) - l_{n,t}(u) & \leq \frac{1}{N} \sum_{n=1}^N \sum_{t=1}^T \langle g_{n,t}, x_{n,t} - u \rangle.
    \end{align*}
    Now, similar to the reward-regret relationship (\Cref{lemma:reward-regret}) and the wealth lower bounds established in~\Cref{lemma:multi_agent_wealth_lb}, we can relate the cumulative wealth to the linear regret.
    Specifically, since~\Cref{eq:wealth_lb_average} holds and the gossip step preserves averages, we set $c_t = -g_t$ and obtain
    \begin{align}
        \frac{1}{N} \sum_{n=1}^N \sum_{t=1}^T \langle g_{n,t}, x_{n,t} \rangle & = - \overline{\w}_T                                  \\
                                                                               & \leq - F_T\br{- \sum_{t=1}^T \bar g_{t}} + \epsilon.
    \end{align}
    Therefore,
    \begin{align}
        \frac{1}{N} & \sum_{n=1}^N \sum_{t=1}^T \langle g_{n,t}, x_{n,t} - u \rangle                                                         \\
                    & \leq - \frac{1}{N} \sum_{n=1}^N \sum_{t=1}^T \langle g_{n,t}, u \rangle - F_T\br{- \sum_{t=1}^T \bar g_{t}} + \epsilon \\
                    & = \left\langle -\sum_{t=1}^T \bar g_{t}, u \right\rangle - F_T\br{- \sum_{t=1}^T \bar g_{t}} + \epsilon                \\
                    & \leq \sup_{\theta \in \Hil{}} \langle \theta, u \rangle - F_T(\theta) + \epsilon                                       \\
                    & = F_T^*(u) + \epsilon.
    \end{align}
\end{proof}
\begin{remark}
    Applying~\cite[Corollary 5]{orabona_coin_2016}, any potential from~\Cref{tab:mappings} satisfies
    \begin{align*}
        R_T^{\text{local}}(u) & \leq \norm{u} \sqrt{T \ln \br{1 + \frac{24T^2 \norm{u}^2}{\epsilon^2}}} + \epsilon,
    \end{align*}
    which gives sublinear regret when $\epsilon = \bigo{\sqrt{T}}$.
\end{remark}
\begin{remark}
    Note that the regret bound in~\Cref{thm:local_avg_regret_bound} only uses the double stochasticity of the gossip matrix $W$, so it holds for any gossip algorithm that preserves averages.
    Notably, this includes the case where agents do not communicate with each other, that is, $W = I_N$.
    This means that \textbf{no matter the network topology}, the average local loss is minimized.
    The effect of the network topology is reflected in the average network regret, which is analyzed in~\Cref{thm:network_loss_regret_bound}.
\end{remark}

We now prove that the network regret is related to the local regret, with a disagreement term that depends on the network topology.
This disagreement will become larger when the output decisions at each node differ more, which in turn is a consequence of the difference between the node losses.
The output decisions are a function of the accumulated subgradient variable at each node.
These are gossiped at each round and their difference is bounded as follows.
\begin{lemma}\label{lemma:G-gossip}
    Let $W$ be a gossip matrix as in~\Cref{def:gossip_matrix}.
    Let $G_{n,t}$ be the accumulated subgradient at Agent $n$, and let $g_t = (g_{1,t}, \ldots, g_{N,t})$ be the vector of subgradients at time $t$.
    If we perform $q(t)$ gossip steps at each round $t$, the error with respect to the average accumulated subgradient at time $t$ evolves as
    \begin{align}
        \norm{G_{n,t} - \bar G_t} \leq \sum_{s=1}^t \rho^{Q(s,t)} \norm{g_{s} - \bar g_s \otimes \mathbf{1}},
    \end{align}
    where $\rho < 1$ is the spectral radius of $W - \frac{1}{N} \mathbf{1} \mathbf{1}^\top$, and
    \begin{equation}
        Q(s,t) = \sum_{r=s}^t q(r).
    \end{equation}
    Since each $g_{n,t} \in (-1,1)$, we can further bound
    \begin{align}
        \norm{G_{n,t} - \bar G_t} \leq \sqrt{N} \sum_{s=1}^t \rho^{Q(s,t)}.
    \end{align}
\end{lemma}
\begin{proof}
    The lemma immediately follows from the triangle inequality and~\Cref{lemma:gossip}.
\end{proof}

\begin{theorem}
    [Network loss regret bound]\label{thm:network_loss_regret_bound}
    Let $\{F_t\}$ be a family of excellent coin-betting potentials as in~\Cref{def:potential}, and let $W$ be a gossip matrix as in~\Cref{def:gossip_matrix}.
    Then, the proposed \algname{}-$\pmb{ii}$ achieves the following regret bound for any comparator $u \in \Hil$:
    \begin{align*}
        R_T^{\text{net}}(u) & = \frac{1}{N} \sum_{n=1}^{N} R_T^{\text{local}}(u)                                                  \\
                            & + \underbrace{2\sqrt{N} \sum_{t=1}^T L_{h_t} \sum_{s=1}^{t-1} \rho^{Q(s,t)}}_{\text{Disagreement}},
    \end{align*}
    where $L_{h_t}$ is the Lipschitz constant of $h_t$ in $[-(t-1),(t-1)]$.
\end{theorem}
\begin{proof}
    Using the definition of network regret in~\Cref{eq:network_regret}, we have:
    \begin{align}
        R_T^{\text{net}}(u) & = \sum_{t=1}^{T} \frac{1}{N} \sum_{n=1}^{N} l_{t}(x_{n,t}) - \sum_{t=1}^{T} l_{t}(u).
    \end{align}
    Next, we add and subtract the average local loss to obtain:
    \begin{align*}
        R_T^{\text{net}}(u) = \sum_{t=1}^{T} \frac{1}{N} \sum_{n=1}^{N} l_{t}(x_{n,t}) & \pm \sum_{t=1}^{T} \frac{1}{N} \sum_{n=1}^{N} l_{n,t}(x_{n,t}) \\
                                                                                       & - \sum_{t=1}^{T} l_{t}(u),
    \end{align*}
    which we can re-arrange to
    \begin{align*}
        R_T^{\text{net}}(u) & = \frac{1}{N} \sum_{n=1}^{N} R_T^{\text{local}}(u)                                                                     \\
                            & + \underbrace{\sum_{t=1}^{T} \frac{1}{N} \sum_{n=1}^{N} \br{l_{t}(x_{n,t}) - l_{n,t}(x_{n,t})}}_{\text{Disagreement}}.
    \end{align*}
    Note that the local regret term can be bounded using~\Cref{thm:local_avg_regret_bound}, so we just have to bound the disagreement term, which we will denote as $D$.
    We can further re-arrange the disagreement term as
    \begin{equation}\label{eq:disagreement}
        D \coloneqq \sum_{t=1}^{T} \frac{1}{N} \sum_{n=1}^{N} \frac{1}{N} \sum_{m=1}^{N} \br{l_{n,t}(x_{m,t}) - l_{n,t}(x_{n,t})}.
    \end{equation}
    Note that we can now use the 1-Lipschitz property to bound each term, that is,
    \begin{align*}
        l_{n,t}(x_{m,t}) - l_{n,t}(x_{n,t}) & \leq \norm{x_{m,t} - x_{n,t}}             \\
                                            & = \norm{h_t(G_{m,t-1}) - h_t(G_{n,t-1})},
    \end{align*}
    where we have used the definition of $x_{m,t}$ according to \algname{}-$\pmb{ii}$.
    Next, we use the property that $h_t$ is $L_{h_t}$-Lipschitz in its domain, and we obtain
    \begin{align}
        \norm{x_{m,t} - x_{n,t}} & \leq L_{h_t} \norm{G_{m,t-1} - G_{n,t-1}}.
    \end{align}
    Now, using the triangle inequality and~\Cref{lemma:G-gossip}, we derive
    \begin{align}
        \norm{x_{m,t} - x_{n,t}} & \leq L_{h_t} \cdot 2\sqrt{N} \sum_{s=1}^{t-1} \rho^{Q(s,t)}.
    \end{align}
    Finally, we plug this back into~\Cref{eq:disagreement} to show
    \begin{align}
        D \leq  2\sqrt{N} \sum_{t=1}^T L_{h_t} \sum_{s=1}^{t-1} \rho^{Q(s,t)}.
    \end{align}
\end{proof}

\begin{lemma}[Square-root disagreement bound]\label{lemma:square-root-disagreement}
    Let $W$ be a gossip matrix as in~\Cref{def:gossip_matrix}, where $\rho$ is the spectral radius of $W - \frac{1}{N} \mathbf{1} \mathbf{1}^\top$, and let
    $\{F_t\}$ be a family of excellent coin-betting potentials, as in~\Cref{def:potential}, where $\{L_{h_t}\}$ are the Lipschitz constants of $\{h_t\}$ derived from $\{F_t\}$ in $[-(t-1), (t-1)]$.
    If we perform $q(t)$ gossip steps at round $t$, such that
    \begin{equation}\label{eq:condition-disagreement}
        \forall t \geq 1, \quad L_{h_t} \sum_{s=1}^{t-1} \rho^{Q(s,t)} \leq C \cdot \frac{1}{\sqrt{t}},
    \end{equation}
    where $Q(s,t) \coloneqq \sum_{r=s}^t q(r)$, we obtain sublinear regret and the disagreement is upper-bounded by $4 C \sqrt{N} \sqrt{T}$.
\end{lemma}
\begin{proof}
    From~\Cref{thm:network_loss_regret_bound}, the disagreement term $D$ is upper bounded as
    \begin{align}
        D \leq 2\sqrt{N} \sum_{t=1}^T L_{h_t} \sum_{s=1}^{t-1} \rho^{Q(s,t)}.
    \end{align}
    Given the condition~\eqref{eq:condition-disagreement}, we can upper bound
    \begin{align}
        D \leq 2\sqrt{N} \sum_{t=1}^T C \cdot \frac{1}{\sqrt{t}}.
    \end{align}
    Note that
    \begin{align}
        \sum_{t=1}^T \frac{1}{\sqrt{t}} \coloneqq H_{T,1/2}
    \end{align}
    is the generalized harmonic number $H_{T,1/2}$.
    From the classic square root reciprocals sum inequality,
    \begin{equation}
        H_{T,1/2} \leq 2 \sqrt{T} - 1,
    \end{equation}
    and, thus, we obtain that
    \begin{equation}
        D \leq 4 C \sqrt{N} \sqrt{T}.
    \end{equation}
\end{proof}

\begin{theorem}[Exponential potentials and sublinear disagreement]\label{thm:exponential-disagreement}
    \algname{}-\pmb{ii}, choosing $\{F_t\}$ to be the exponential potentials from~\Cref{tab:mappings}, defined as
    \begin{equation}
        F_t(x) = \frac{\epsilon}{\sqrt{t}} \exp\br{\frac{x^2}{2t}},
    \end{equation}
    and choosing the gossip step function to be
    \begin{equation}
        q(t) = \lceil c \cdot t \rceil,
    \end{equation}
    for any $c \geq \frac{-3}{2\ln \rho}$ has sublinear regret and disagreement.
\end{theorem}
\begin{proof}
    The betting function derived from the exponential potential is
    \begin{align}
        h_t(x) = \tanh \br{\frac{x}{t}} \cdot F_{t-1}(x),
    \end{align}
    which is odd and has derivative
    \begin{align}
        h_t'(x) = F_t(x) \frac{\sech^2{\frac{x}{t}} + x \tanh{\frac{x}{t}}}{t}.
    \end{align}
    For $t\geq 1$, and $|x| \leq t-1$, we can derive that
    \begin{align}
        |h_t'(x)| & \leq  F_t(t-1) \br{1 - \frac{2}{1 + e^2}}                                                        \\
                  & \leq \frac{\epsilon}{\sqrt{t}} \exp \br{\frac{\br{t-1}^2}{2t}} \cdot \br{1 - \frac{2}{1 + e^2}},
    \end{align}
    and we can use this quantity as $L_{h_t}$, needed in~\Cref{lemma:square-root-disagreement}.
    The proof is complete if~\Cref{lemma:square-root-disagreement} holds and we apply it with $C \coloneqq \epsilon \cdot \br{1 - \frac{2}{1 + e^2}}$.
    To use~\Cref{lemma:square-root-disagreement}, we need to show the validity of Ineq.~\eqref{eq:condition-disagreement}, i.e.,
    \begin{align*}
        \forall t \geq 1, \quad \exp \br{\frac{\br{t-1}^2}{2t}} \sum_{s=1}^{t-1} \rho^{Q(s,t)} \leq 1.
    \end{align*}
    For $t=1$, Ineq.~\eqref{eq:condition-disagreement} is trivial.
    Next, for $t \geq 2$, we can ensure that
    \begin{align*}
        \exp \br{\frac{\br{t-1}^2}{2t}} \sum_{s=1}^{t-1} \rho^{Q(s,t)} \leq \exp \br{\frac{t}{2}} \sum_{s=1}^{t-1} \rho^{Q(s,t)},
    \end{align*}
    so the condition is satisfied if
    \begin{align*}
        e^{t/2}\sum_{s=1}^{t-1} \rho^{Q(s,t)} \leq 1.
    \end{align*}
    Now, let $q(t) = c \cdot t$.
    In practice, we can take the ceiling of this quantity, since the number of gossip iterations has to be an integer.
    For ease of analysis, we will disregard the ceiling, but the logic follows analogously.
    Notice that we can express
    \begin{align}
        Q(s,t) = c \cdot \frac{(t-s+1)(s+t)}{2},
    \end{align}
    which is smallest when $s=t-1$.
    Thus, we can upper bound
    \begin{align*}
        e^{t/2} \sum_{s=1}^{t-1} \rho^{Q(s,t)} \leq e^{t/2} (t-1) \rho^{c(2t-1)}.
    \end{align*}
    We want to ensure that
    \begin{align*}
        e^{t/2} (t-1) \rho^{c(2t-1)} \leq 1
    \end{align*}
    for all $t \geq 2$.
    Taking logarithms, we obtain
    \begin{align}
        \frac{t}{2} + \ln(t-1) + \frac{c\ln \rho}{2}(2t-1) \leq 0 \label{eq:taking-logs}
    \end{align}
    Notice that if the LHS is non-increasing, we only need to verify Ineq.~\eqref{eq:taking-logs} at $t=2$.
    The derivative of the LHS is
    \begin{align*}
        \frac{1}{2} + \frac{1}{t-1} + c\ln(\rho),
    \end{align*}
    and ensuring
    \begin{equation}
        c \geq \frac{-3}{2\ln(\rho)}, \label{eq:cond_monotonicity}
    \end{equation}
    it is non-positive for all $t \geq 2$, which ensures the non-increasing property.
    If Ineq.~\eqref{eq:cond_monotonicity} is satisfied, we can substitute $t=2$ in Ineq.~\eqref{eq:taking-logs} and verify that the bound holds, which concludes our proof.
\end{proof}
\begin{theorem}[KT potentials and sublinear disagreement]\label{thm:KT-disagreement}
    \algname{}-\pmb{ii}, choosing $\{F_t\}$ to be the KT potentials from~\Cref{tab:mappings}, defined as
    \begin{equation}
        F_t(x) = \epsilon \frac{2^t \cdot \Gamma\br{\frac{t+1}{2} + \frac{x}{2}} \cdot \Gamma\br{\frac{t+1}{2} - \frac{x}{2}}}{\pi \cdot t!},
    \end{equation}
    and choosing the gossip step function to be
    \begin{equation}
        q(t) = \lceil c \cdot t \rceil,
    \end{equation}
    for any $c \geq \frac{-2\ln(2)}{\ln(\rho)}$ has sublinear regret and disagreement.
\end{theorem}
\begin{proof}
    The betting function derived from the KT potential is
    \begin{align}
        h_t(x) = \frac{x}{t} \cdot F_{t-1}(x).
    \end{align}
    Following a similar procedure to the one in~\Cref{thm:exponential-disagreement}, we obtain that for $t \geq 1$, and $|x| \leq t-1$, $|h'_t|$ is upper bounded by
    \begin{align}
        \frac{2^{t-2}(\ln t - \psi(1/2))}{\sqrt{\pi t}},
    \end{align}
    where $\psi$ is the digamma function.
    Thus, we can set $C\coloneqq 4\sqrt{\pi}$, and apply~\Cref{lemma:square-root-disagreement}.
    For a valid~\Cref{lemma:square-root-disagreement}, we need to verify Ineq.~\eqref{eq:condition-disagreement}, which in this case means ensuring
    \begin{align}
        \forall t \geq 1, \quad 2^{t}(\ln t - \psi(1/2)) \sum_{s=1}^{t-1} \rho^{Q(s,t)} \leq 1.
    \end{align}
    We set $q(t) = c \cdot t$ and check that
    \begin{align}
        2^{t}(\ln t - \psi(1/2)) (t-1) \rho^{c(2t-1)} \leq 1.
    \end{align}
    For $t=1$, this is trivial.
    For $t \geq 2$, we can take logarithms and obtain
    \begin{align*}
        t\ln(2) + \ln(\ln t - \psi(1/2)) + \ln(t-1) + \frac{c\ln \rho}{2}(2t-1) \leq 0.
    \end{align*}
    Next, we use the bound
    \begin{align*}
        t\ln(2) + \ln(\ln t - \psi(1/2)) + \ln(t-1) \leq 2 t \ln(2)
    \end{align*}
    for all $t \geq 2$, and the condition simplifies to
    \begin{align*}
        2 t \ln(2) + \frac{c\ln \rho}{2}(2t-1) \leq 0,
    \end{align*}
    which is satisfied when $c \geq \frac{-2\ln(2)}{\ln(\rho)}$.
    This concludes our proof.
\end{proof}

\begin{remark}
    \Cref{thm:exponential-disagreement,thm:KT-disagreement} provide a clear trade-off between communication and performance.
    They establish that by performing $\bigo{t}$ gossip steps at each round, a communication cost that grows linearly with time, the disagreement term can be controlled to be at most in the order of the local regret term, thus ensuring sublinear network regret.
\end{remark}

\section{Operational Details}\label{sec:operational-details}
To bridge the gap between our theoretical analysis and practical implementation, this section details the operational flow of \algname{}.
We illustrate how the algorithm operates in a fully decentralized manner using an online linear regression task, which is common in applications like distributed sensing, as an example.

Consider a weather sensor network deployed in a large agricultural field.
Each sensor acts as an agent, forming a wireless mesh network with nearby sensors.
The collective goal is to learn a single robust model to predict a value, such as air temperature or soil moisture, based on other local readings.

At each time step $t$, every sensor $n$ makes a prediction using its current model parameters $x_{n,t}$ and a vector of local features $z_{n,t}$ (e.g., \texttt{[humidity, pressure]}).
After observing the true value of the temperature or soil moisture, $y_{n,t}$, it computes its loss.
For this example, we use the absolute loss
\begin{equation}
    \ell_{n,t}(x) = | \langle x, z_{n,t} \rangle - y_{n,t} |,\label{eq:absolute_loss}
\end{equation}
with subgradient
\begin{equation}
    g_{n,t} = \text{sign}(\langle x, z_{n,t} \rangle - y_{n,t}) \cdot z_{n,t}.\label{eq:absolute_subgradient}
\end{equation}
The sensors then use this subgradient to update their internal state and communicate with neighbors to collaboratively refine their models.

The operational flow at each agent is a four-step loop.
First, the agent makes a \textbf{prediction} by computing its decision $x_{n,t}$ from its current internal state.
Following this, it \textbf{observes} feedback from the environment (e.g., by measuring the true temperature or soil moisture) and computes the local subgradient $g_{n,t}$.
This subgradient is then used to perform a \textbf{local update} on the agent's internal state, creating a temporary version.
The round concludes by \textbf{communicating} this temporary state with the agent's direct neighbors through a gossip protocol, averaging their information to produce the final state for the next round.

The key to decentralization lies in the communication step.
This is a multi-round process where information propagates across the network.
For $q(t)$ rounds of gossip, each agent $n$ only exchanges its current state with its immediate neighbors $\mathcal{N}(n)$.
The state is updated using a weighted average, with weights defined by the gossip matrix $W$.
These weights can be constructed locally using, for example, the Metropolis-Hastings rule, which only requires an agent to know its own degree and the degrees of its neighbors:
\begin{equation}\label{eq:Metropolis-Hastings}
    W_{mn} = \begin{cases}
        1 - \sum_{j \in \mathcal{N}(n) \setminus \{n\}} W_{jn}, & \text{if } m = n,                \\
        \frac{1}{\max\{d_n, d_m\} + 1},                         & \text{if } m \in \mathcal{N}(n), \\
        0,                                                      & \text{otherwise}.
    \end{cases}
\end{equation}
After $q(t)$ local exchanges, the agent has its final state for round $t$, ready for the next prediction.

In practice, the two variants of \algname{} differ in what constitutes the agent's state and how the decision is computed, which in turn affects the local update and communication steps.

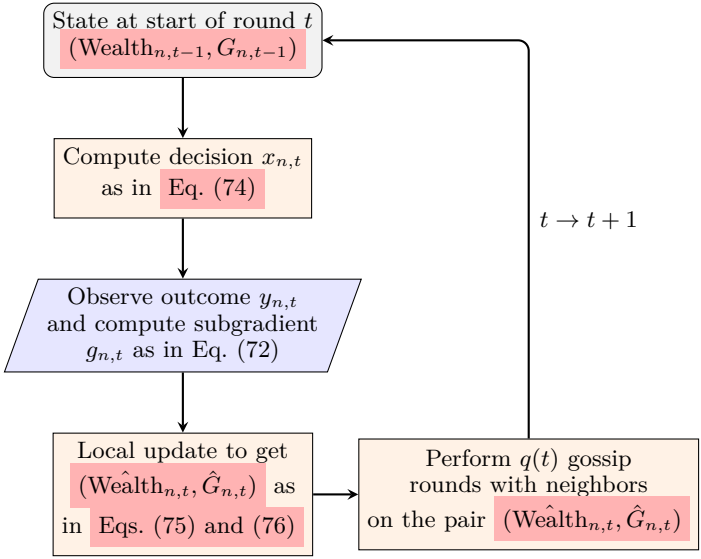
\begin{figure}[tbp!]
    \centering
    \begin{tikzpicture}[
            node distance=0.8cm and .5cm,
            font=\small
        ]
        \node (state) [state] {State at start of round $t$ \\ \colorbox{red!30}{$(\w_{n,t-1}, G_{n,t-1})$}};
        \node (compute) [process, below=of state] {Compute decision $x_{n,t}$ \\ as in~\colorbox{red!30}{\Cref{eq:deco1_bet_vec}}};
        \node (observe) [io, below=of compute] {Observe outcome $y_{n,t}$ \\
            and compute subgradient \\
            $g_{n,t}$ as in~\Cref{eq:absolute_subgradient}};
        \node (update) [process, below=of observe] {Local update to get \\\colorbox{red!30}{$(\hat{\w}_{n,t}, \hat{G}_{n,t})$} as\\
            in~\colorbox{red!30}{\Cref{eq:local-update-wealth,eq:local-update-G}}};
        \node (gossip) [process, right=of update, xshift=+.1cm] {Perform $q(t)$ gossip \\
            rounds with neighbors \\
            on the pair \colorbox{red!30}{$(\hat{\w}_{n,t}, \hat{G}_{n,t})$}};

        \draw [arrow] (state) -- (compute);
        \draw [arrow] (compute) -- (observe);
        \draw [arrow] (observe) -- (update);
        \draw [arrow] (update) -- (gossip);
        \draw [arrow] (gossip) |- node[above, pos=0.25, xshift=.8cm] {$t \to t+1$} (state);
    \end{tikzpicture}
    \caption{Operational loop for an agent using \algname{}-\pmb{i}.
        Key elements specific to this variant, related to tracking and communicating wealth, are highlighted in red.}\label{fig:flow-i}
\end{figure}

In~\Cref{fig:flow-i}, we illustrate the operational loop for \algname{}-\pmb{i}, which directly adapts the standard single-agent coin-betting algorithm.
In this variant, an agent's state consists of its wealth and accumulated gradients, $(\w_{n,t-1}, G_{n,t-1})$.
The prediction is computed from this pair.
For the exponential potential, it is
\begin{equation}
    x_{n,t} = \tanh\br{\frac{\norm{G_{n,t-1}}}{t}} \cdot \w_{n,t-1} \cdot \frac{G_{n,t-1}}{\norm{G_{n,t-1}}}.\label{eq:deco1_bet_vec}
\end{equation}
The local update then affects both state components:
\begin{align}
    \hat{\w}_{n,t} & = \w_{n,t-1} - \langle g_{n,t}, x_{n,t} \rangle,\label{eq:local-update-wealth} \\
    \hat{G}_{n,t}  & = G_{n,t-1} + g_{n,t}.\label{eq:local-update-G}
\end{align}
For communication, the agent gossips the pair $(\hat{\w}_{n,t}, \hat{G}_{n,t})$.
Although this increases the communication load, the wealth is a scalar, which is often negligible compared to the dimension of the gradient vector.

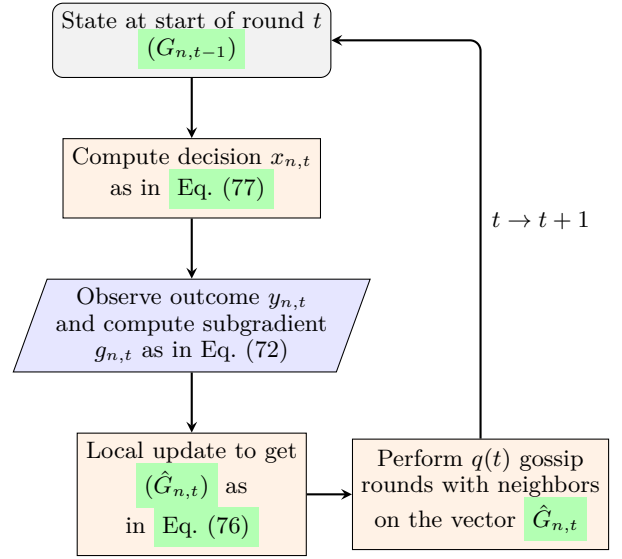
\begin{figure}[tbp!]
    \centering
    \begin{tikzpicture}[
            node distance=0.8cm and .5cm,
            font=\small
        ]
        \node (state) [state] {State at start of round $t$ \\ \colorbox{green!30}{$(G_{n,t-1})$}};
        \node (compute) [process, below=of state] {Compute decision $x_{n,t}$ \\ as in~\colorbox{green!30}{\Cref{eq:deco2_bet_vec}}};
        \node (observe) [io, below=of compute] {Observe outcome $y_{n,t}$ \\
            and compute subgradient \\
            $g_{n,t}$ as in~\Cref{eq:absolute_subgradient}};
        \node (update) [process, below=of observe] {Local update to get \\\colorbox{green!30}{$(\hat{G}_{n,t})$} as\\
            in~\colorbox{green!30}{\Cref{eq:local-update-G}}};
        \node (gossip) [process, right=of update, xshift=+.1cm] {Perform $q(t)$ gossip \\
            rounds with neighbors\\
            on the vector \colorbox{green!30}{$\hat{G}_{n,t}$}};

        \draw [arrow] (state) -- (compute);
        \draw [arrow] (compute) -- (observe);
        \draw [arrow] (observe) -- (update);
        \draw [arrow] (update) -- (gossip);
        \draw [arrow] (gossip) |- node[above, pos=0.25, xshift=.8cm] {$t \to t+1$} (state);
    \end{tikzpicture}
    \caption{Operational loop for \algname{}-\pmb{ii}.
        The process is simplified by decoupling from wealth, requiring only the accumulated gradients to be maintained and gossiped.
        Differences from \algname{}-\pmb{i} are highlighted in green.}\label{fig:flow-ii}
\end{figure}

\Cref{fig:flow-ii} depicts the operational loop for \algname{}-\pmb{ii}, which is designed to be more communication-efficient and analytically tractable.
However, the performance of \algname{}-\pmb{ii} is often worse than that of \algname{}-\pmb{i}.
Here, the agent's state only consists of the accumulated gradients $G_{n,t-1}$.
The prediction is computed directly from this state using the betting function $h_t(\cdot)$.
For the exponential potential, this is
\begin{equation}
    x_{n,t} = \underbrace{\tanh\br{\frac{\norm{G_{n,t-1}}}{t}} \cdot \frac{\epsilon e^{\frac{\norm{G_{n,t-1}}^2}{2(t-1)}}}{\sqrt{t-1}}}_{h_t(\norm{G_{n,t-1}})} \cdot \frac{G_{n,t-1}}{\norm{G_{n,t-1}}}.\label{eq:deco2_bet_vec}
\end{equation}
Then, the local update is only applied to the accumulated gradient vector as in~\Cref{eq:local-update-G}.
Thus, the agent only needs to gossip the vector $\hat{G}_{n,t}$.

\section{Experiments}\label{sec:experiments}

To validate our theoretical findings, we conduct a series of experiments on both synthetic and real-world datasets.
Our implementation is publicly available.\footnote{Code available at: \url{https://github.com/TomasOrtega/Deco}}
All experiments compare our decentralized coin-betting algorithms, which we will refer to as \textbf{\algname{}-i} and \textbf{\algname{}-ii}, against standard baselines.

\subsection{Experimental setup}

\subsubsection*{Algorithms compared}
We evaluate the performance of our two proposed variants from \algname{}, with two different potential functions from \Cref{tab:mappings}.
Therefore, we have four distinct parameter-free algorithms:
\begin{itemize}
    \item \textbf{\algname{}-i (exp)}: Version (\pmb{i}) using the exponential potential, generating a $\tanh$ betting fraction.
    \item \textbf{\algname{}-ii (exp)}: Version (\pmb{ii}) using the exponential potential.
    \item \textbf{\algname{}-i (KT)}: Version (\pmb{i}) using the KT potential, based on the Krichevsky-Trofimov bettor, which produces a linear betting fraction, $\beta_t(x) = x/t$.
    \item \textbf{\algname{}-ii (KT)}: Version (\pmb{ii}) using the KT potential.
\end{itemize}
We compare them against two baselines:
\begin{itemize}
    \item \textbf{Decentralized Online Gradient Descent (DOGD)}: A standard decentralized algorithm where each agent performs a gradient step followed by a gossip step.
          The learning rate follows a decreasing schedule $\eta_t = \eta_0 / \sqrt{t}$.
    \item \textbf{Centralized}: An idealized oracle in which a single coin-betting agent (using the KT potential) has access to the average gradient $\bar{g}_t = \frac{1}{N}\sum_{n=1}^N g_{n,t}$ at each round.
\end{itemize}

\subsubsection*{Network and evaluation}
Unless otherwise specified, all experiments are conducted in a network of $N=20$ agents.
Performance is measured by \textbf{cumulative network loss}, defined as $\sum_{t=1}^{T} \frac{1}{N} \sum_{n=1}^{N} l_{t}(x_{n,t})$.
For the gossip step, we use a single round of communication ($q(t)=1$) by default to represent a communication-constrained scenario.

\subsubsection*{Datasets}
We use both a synthetic dataset for controlled experiments and several real-world datasets to test for robustness.
\begin{itemize}
    \item \textbf{Synthetic data}: We simulate an online linear regression task.
          First, a true parameter vector $u^* \in \mathbb{R}^{10}$ is drawn from a standard normal distribution and fixed.
          To model data heterogeneity, each agent $n$ is assigned a unique feature center $\mu_n \in \mathbb{R}^{10}$, drawn from a normal distribution.
          At each round $t$, a common base feature vector $z'_t$ is drawn from a standard normal distribution.
          Agent $n$ then observes a personalized feature vector $z_{n,t} = z'_t + \mu_n$, which is subsequently normalized to have a unit $\ell_2$-norm ($\|z_{n,t}\|_2 = 1$).
          Each agent $n$ observes a label $y_{n,t} = \langle u^*, z_{n,t} \rangle + \nu_{n,t}$, where $\nu_{n,t}$ is a small Gaussian noise term.
          The loss function is the absolute loss, $\ell_{n,t}(x) = |\langle x, z_{n,t} \rangle - y_{n,t}|$.
    \item \textbf{Real-world data}: We use four regression datasets from the LIBSVM library: \texttt{cadata}, \texttt{cpusmall}, \texttt{space\_ga}, and \texttt{abalone}~\cite{chang_libsvm_2011}.
          The features of these datasets are pre-normalized.
\end{itemize}

\subsubsection*{Implementation note}
A notable challenge while computing quantities involving factorial or Gamma functions, such as in the KT potential shown in~\Cref{tab:mappings}, is numerical stability.
Direct computation of the Gamma function, $\Gamma(z)$, can easily lead to numerical overflow for even moderately large arguments.

To address this, the KT potential can be expressed in a more stable form using the Beta function, $B(a,b) = \frac{\Gamma(a)\Gamma(b)}{\Gamma(a+b)}$.
Namely,
\begin{equation}
    F_t(x) = \frac{\epsilon \cdot 2^t}{\pi} B\br{\frac{t+1+|x|}{2}, \frac{t+1-|x|}{2}}.
\end{equation}
However, even the Beta function can underflow to zero.
Our implementation avoids these issues by performing calculations in the log-domain:
\begin{equation}
    \ln F_t(x) = \ln\br{\epsilon / \pi} + t \ln(2) + \ln B\br{a, b}.
\end{equation}
The term $\ln B(a,b)$ is computed using a highly optimized log-beta function, Python's \texttt{scipy.special.betaln}, which sidesteps the intermediate calculation of large or small numbers.
The final value of $F_t(x)$ is then recovered by exponentiating the result.

\subsection{The challenge of hyperparameter tuning}

\begin{figure}[tbp!]
    \centering
    \includegraphics[width=\columnwidth]{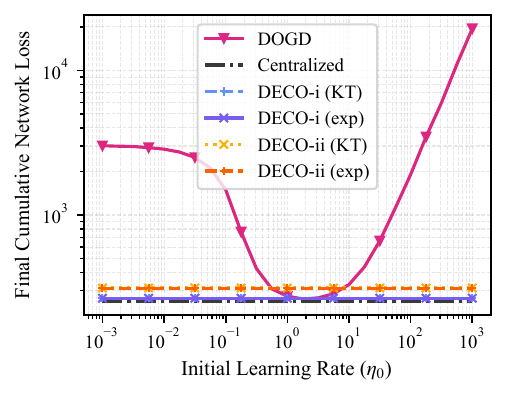}
    \caption{Final cumulative network loss of DOGD as a function of its initial learning rate ($\eta_0$) on synthetic data, compared to the constant performance of our parameter-free \algname{} algorithms.
        The U-shaped curve for DOGD highlights its extreme sensitivity to hyperparameter choice, where a suboptimal setting leads to loss that is orders of magnitude worse.
        Our \algname{} algorithms achieve good performance without any tuning, regardless of the potential function used.}\label{fig:dgd_sensitivity}
\end{figure}

A primary motivation for our work is the brittleness of traditional algorithms that require manual tuning.
To illustrate this, we first conduct an experiment focusing on the sensitivity of DOGD to its initial learning rate, $\eta_0$.
We run the DOGD algorithm on our synthetic dataset over a wide range of learning rates, from $10^{-3}$ to $10^3$, and plot the final cumulative network loss after $T=3000$ iterations.

As shown in \Cref{fig:dgd_sensitivity}, the performance of DOGD highly depends on $\eta_0$.
The loss forms a sharp ``U'' shape, indicating that only a narrow range of values leads to good performance.
In contrast, the performance of all four \algname{} variants is represented by flat horizontal lines, as they do not require such tuning.
This plot showcases the advantage of a parameter-free approach, which reliably achieves low loss.

Since KT and exponential potentials provide very similar performance, for clarity of the plots, we only show the centralized results using the KT potential.
Also, since the difference between the two potential functions for \algname{} is small, in what follows, we only focus on results using the KT potential.

\subsection{Impact of network connectivity}
Our theoretical analysis in \Cref{thm:network_loss_regret_bound} shows that network regret is influenced by a disagreement term, which should decrease as network connectivity improves.
To test this empirically, we evaluate the \algname{}-ii (KT) algorithm on several Erd\H{o}s-R\'enyi (ER) random graphs~\cite{erdos_random_1959} with different edge probabilities, $p$.
A higher $p$ results in a denser graph with a smaller spectral radius $\rho$, facilitating faster information mixing.

\begin{figure}[tbp!]
    \centering
    \includegraphics[width=\columnwidth]{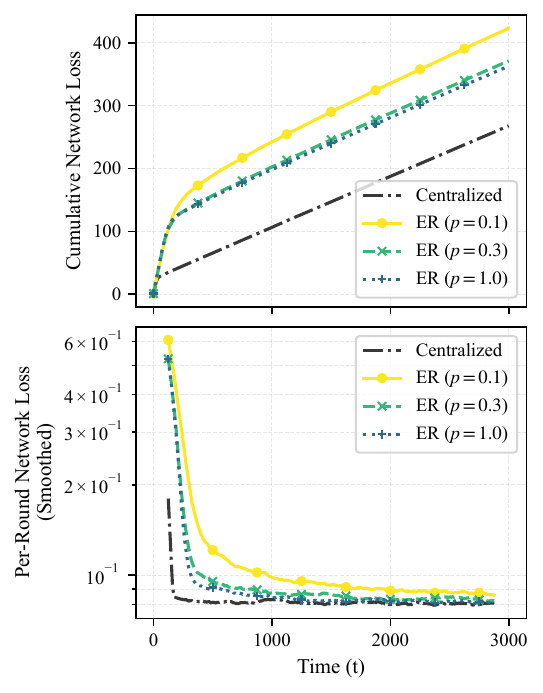}
    \caption{Performance of \algname{}-ii (KT) on synthetic data for Erd\H{o}s-R\'enyi random graphs with varying edge probability $p$.
        \textbf{Top}: Cumulative network loss.
        Cumulative loss growth improves with better connectivity, as faster information mixing reduces the disagreement among agents.
        \textbf{Bottom}: Per-round network loss (smoothed with a 250-round moving average).
        The smoothed loss shows that more connected networks not only accumulate less total loss but also converge faster to the centralized ideal's steady-state per-round loss.}\label{fig:connectivity_impact}
\end{figure}

The results are presented in \Cref{fig:connectivity_impact}.
The top panel shows that, as predicted by the theory, the cumulative network loss decreases with better connectivity.
On the sparsely connected graph ($p=0.1$), the cumulative loss is highest due to slower consensus.
Note that there is an initial jump that results in the deviation of the gossiped algorithms from the centralized oracle, which is due to the lag time of the gossiped algorithms in achieving consensus.
As connectivity increases ($p=0.3$ and $p=1.0$), the growth of the cumulative loss curves approaches that of the centralized oracle, demonstrating that improved communication directly mitigates the cost of decentralization.

The bottom panel, which displays the smoothed per-round network loss, provides further insight.
It illustrates that not only do more connected networks accumulate less total loss, but they also converge faster to the steady-state loss rate of the centralized ideal.

\subsection{The communication-regret trade-off}
\begin{figure}[tbp!]
    \centering
    \includegraphics[width=\columnwidth]{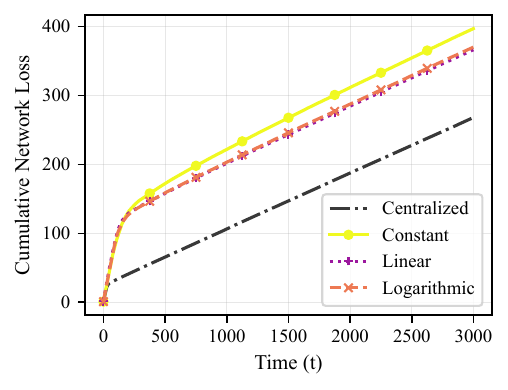}
    \caption{Cumulative network loss of \algname{}-ii (KT) on a cycle graph with different gossip schedules, $q(t)$.
        Increasing the communication budget from constant to logarithmic and linear progressively reduces loss, allowing the algorithm to approach the centralized oracle's performance.
        This highlights the explicit trade-off between communication cost and learning performance.}\label{fig:gossip_tradeoff}
\end{figure}

Our theory suggests that performing more gossip steps per round, $q(t)$, can further reduce disagreement and improve regret, at the cost of increased communication.
We investigate this trade-off by running \algname{}-ii (KT) on a cycle graph with three different gossip schedules:
\begin{itemize}
    \item \textbf{Constant}: $q(t) = 1$ (minimal communication).
    \item \textbf{Logarithmic}: $q(t) = \lceil \ln(t+1) \rceil$.
    \item \textbf{Linear}: $q(t) = \lceil 0.1 \cdot t \rceil$.
\end{itemize}
\Cref{fig:gossip_tradeoff} shows a clear trade-off between communication and performance.
While a single gossip step already performs well, increasing the number of communications over time systematically lowers the cumulative loss curves.
A linear schedule, as suggested by our theory, allows the decentralized algorithm to closely track the performance of the centralized oracle.

\subsection{Performance on real-world datasets}

\begin{figure*}[tbp!]
    \centering
    \includegraphics[width=\textwidth]{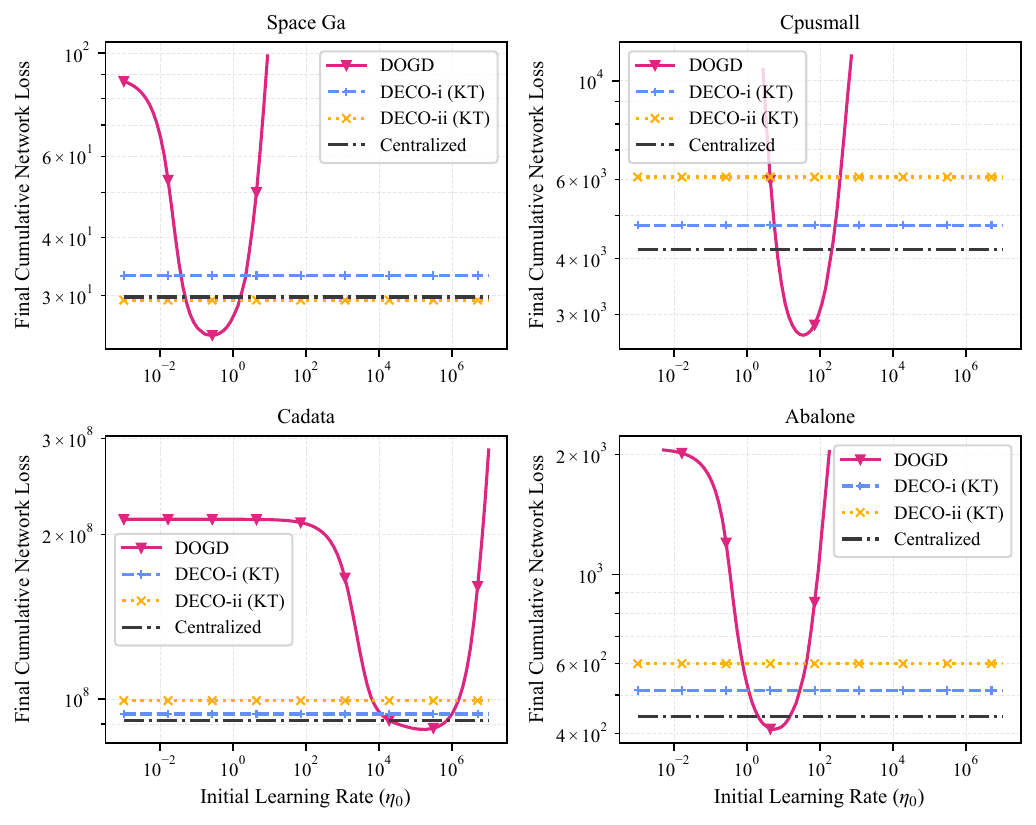}
    \caption{Final cumulative network loss on four real-world LIBSVM datasets.
        Each panel shows the sensitivity of DOGD to its initial learning rate ($\eta_0$), producing a characteristic U-shaped curve.
        Note that we do not plot loss values that are five times the minimum value to allow for good visualization.
        The horizontal lines show the performance of the parameter-free \algname{} variants, which achieve competitive loss without any tuning.
        This highlights the unreliability of DOGD when its hyperparameters are not properly set for a specific dataset.}\label{fig:real_data_results}
\end{figure*}

Finally, to assess robustness, we evaluate all variants of the algorithm on four real-world LIBSVM regression datasets.
These experiments are run on a cycle graph with a constant gossip schedule ($q(t)=1$).
For the DOGD baseline, we perform a hyperparameter sweep over a wider range of initial learning rates, from $10^{-3}$ to $10^7$, to account for the diverse characteristics of these datasets.

The results in \Cref{fig:real_data_results} reinforce our findings.
For each dataset, DOGD's performance exhibits a sharp U-shaped dependence on its initial learning rate $\eta_0$.
Note that we do not plot loss values that are five times the minimum value to allow good visualization.
The optimal value for $\eta_0$ varies significantly from one dataset to another, making it impossible to choose a single reliable setting.
In contrast, the \algname{} variants (using the KT potential) consistently achieve low cumulative loss across all datasets without requiring any parameter selection.
This shows that our framework is applicable to the diverse characteristics of real-world data.

\section{Conclusion}\label{sec:conclusion}

This paper introduced a family of decentralized, parameter-free online learning algorithms, termed \algname{}, designed to address the challenge of hyperparameter tuning in multi-agent systems.
By combining the comparator-adaptive mechanism of coin-betting with information mixing via gossip protocols, our approach removes the need for setting a learning rate, a significant practical barrier for standard methods like Decentralized Online Gradient Descent (DOGD).
A novel betting function formulation was also proposed to facilitate the theoretical analysis in the multi-agent context.

We provided a theoretical analysis showing that our algorithms achieve sublinear network regret.
The regret bounds characterize the influence of network connectivity and communication effort through a disagreement term.
Theoretically, a key open question is whether a constant number of gossip steps per round, $q(t)=\mathcal{O}(1)$, is sufficient to guarantee sublinear network regret.
We conjecture that this is the case.
Our theoretical findings were supported by a series of experiments on both synthetic and real-world datasets.
The results demonstrated that \algname{} achieves performance comparable to a well-tuned DOGD baseline across various datasets and network topologies without requiring any parameter selection.

\bibliographystyle{ieeetr}
\bibliography{zotero}
\end{document}

%% file: decentralized_algo.tex
\begin{algorithmic}[1]
    \STATE{} \textbf{Initialize:} For each agent $n=1, \dots, N$:
    \STATE{} \hspace{\algorithmicindent} $G_{n,0} = \mathbf{0}$.
    \STATE{} \hspace{\algorithmicindent} \colorbox{red!30}{(\pmb{i}) $\w_{n,0} = \epsilon > 0$.}
    \STATE{} \textbf{for} $t=1, 2, \dots, T$ \textbf{do}
    \STATE{} \hspace{\algorithmicindent} \textbf{for each agent} $n=1, \dots, N$ \textbf{in parallel do}
    \STATE{} \hspace{\algorithmicindent}\hspace{\algorithmicindent} \colorbox{red!30}{(\pmb{i}) Compute bet:}
    \begin{center}
        \colorbox{red!30}{$x_{n,t} = \beta_t(G_{n, t-1}) \cdot \w_{n, t-1}$.}
    \end{center}
    \STATE{} \hspace{\algorithmicindent}\hspace{\algorithmicindent} \colorbox{green!30}{(\pmb{ii}) Compute bet using betting function:}
    \begin{center}
        \colorbox{green!30}{$x_{n,t} = h_t(G_{n, t-1})$.}
    \end{center}
    \STATE{} \hspace{\algorithmicindent}\hspace{\algorithmicindent} Play $x_{n,t}$ and receive subgradient $g_{n,t}$.
    \STATE{} \hspace{\algorithmicindent}\hspace{\algorithmicindent} Update local accumulated gradient: \\
    \begin{center}
        $\hat{G}_{n,t} = G_{n,t-1} - g_{n,t}$.
    \end{center}
    \STATE{} \hspace{\algorithmicindent}\hspace{\algorithmicindent} \colorbox{red!30}{(\pmb{i}) Update local wealth:} \\
    \begin{center}
        \colorbox{red!30}{$\hat{\w}_{n,t} = \w_{n,t-1} - \langle g_{n,t}, x_{n,t} \rangle$.}
    \end{center}
    \STATE{} \hspace{\algorithmicindent} \textbf{end for}
    \STATE{} \hspace{\algorithmicindent} \textbf{Gossip Step:} Perform $q(t)$ rounds of gossip.
    \STATE{} \hspace{\algorithmicindent} \textbf{for each agent} $n=1, \dots, N$ \textbf{in parallel do}
    \STATE{} \hspace{\algorithmicindent}\hspace{\algorithmicindent} $G_{n,t} = \sum_{m=1}^N {(W^{q(t)})}_{nm} \hat{G}_{m,t}$.
    \STATE{} \hspace{\algorithmicindent}\hspace{\algorithmicindent} \colorbox{red!30}{(\pmb{i}) $\w_{n,t} = \sum_{m=1}^N {(W^{q(t)})}_{nm} \hat{\w}_{m,t}$.}
    \STATE{} \hspace{\algorithmicindent} \textbf{end for}
    \STATE{} \textbf{end for}
\end{algorithmic}